\documentclass[conference]{IEEEtran}
\IEEEoverridecommandlockouts
\usepackage{cite}
\usepackage{amsmath,amssymb,amsfonts}
\usepackage{algorithmic}
\usepackage{graphicx}
\usepackage{textcomp}
\usepackage{xcolor}
\usepackage{xspace}
\usepackage{subcaption}
\usepackage{tikz}
\def\BibTeX{{\rm B\kern-.05em{\sc i\kern-.025em b}\kern-.08em
    T\kern-.1667em\lower.7ex\hbox{E}\kern-.125emX}}

\newcommand{\FLC}{FLSea\xspace}

\usepackage{hyperref}
\usepackage{wrapfig}
\usepackage{soul}

\begin{document}

\title{ Self-Supervised Monocular Depth Underwater 
\thanks{The research was funded by Israel Science Foundation grant $\#680/18$, the Israeli Ministry of Science and Technology grant $\#3-15621$, the Israel Data Science Initiative (IDSI) of the Council for Higher Education in Israel, the Data Science Research Center at the University of Haifa, and the European Union’s Horizon 2020 research and innovation programme under grant agreement No. GA 101016958.}
}

\author{\IEEEauthorblockN{Shlomi Amitai, Itzik Klein~\IEEEmembership{Senior Member,~IEEE}, and Tali Treibitz}
\IEEEauthorblockA{The Hatter Department of Marine Technologies \\
Charney School of Marine Sciences, University of Haifa, Haifa, Israel\\
shlomi.amitai@gmail.com, \{kitzik,ttreibitz\}@univ.haifa.ac.il}
}

\maketitle

\begin{abstract}
Depth estimation is critical for any robotic system. In the past years estimation of  depth from monocular images have shown great improvement, however, in the underwater environment results are still lagging behind due to appearance changes caused by the medium. So far little effort has been invested on overcoming this.
Moreover, underwater, there are more limitations for using high resolution depth sensors, this makes generating ground truth for learning methods another enormous obstacle. So far unsupervised methods that tried to solve this have achieved very limited success as they relied on domain transfer from dataset in air. We suggest training  using subsequent frames self-supervised by a reprojection loss, as was demonstrated successfully above water. We suggest several additions to the self-supervised framework to cope with the underwater environment and achieve state-of-the-art results on a challenging forward-looking underwater dataset. 
\end{abstract}

\section{Introduction}

\begin{figure}[t]
\centerline{\includegraphics[width=\linewidth]{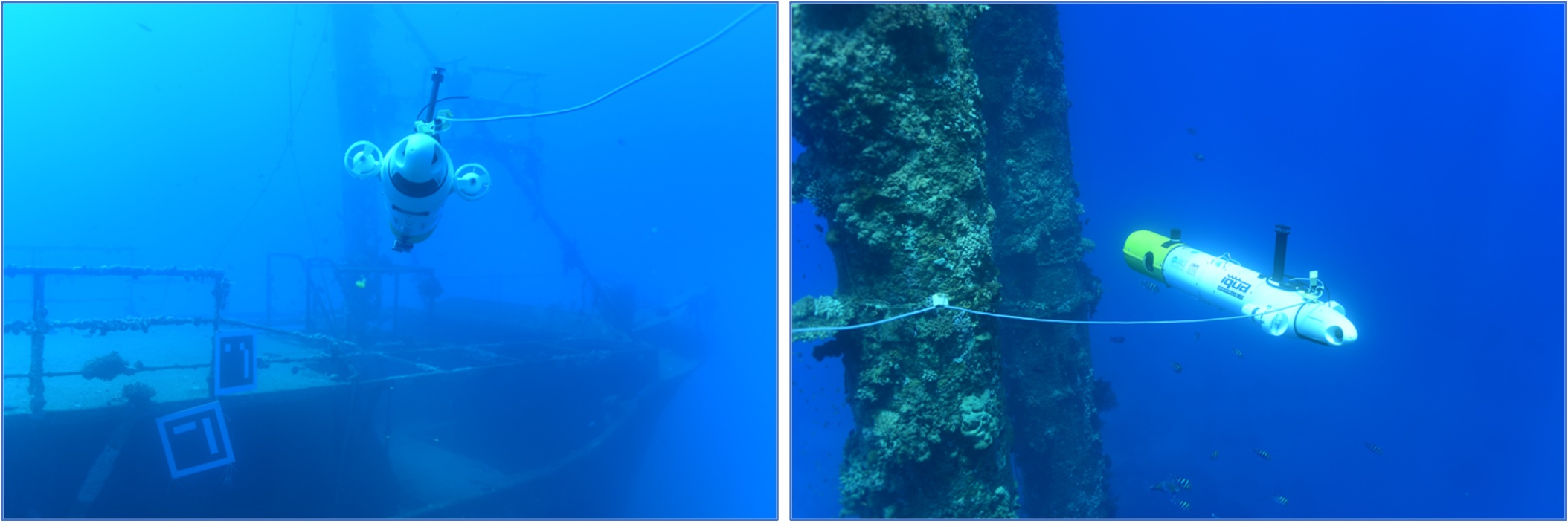}}
\caption{The ALICE autonomous underwater vehicle (AUV)~\cite{gutnik2022adaptation} facing obstacles. Monocular depth maps can aid obstacle avoidance and decision making. }
\label{fig:robot}
\end{figure}

There is a  wide range of target applications for depth estimation,  from obstacle detection to object measurement and from 3D reconstruction to image enhancement. 
Underwater depth estimation (note that here depth refers to the object range, and not to the depth under water) is important for Autonomous Underwater Vehicles (AUVs)~\cite{gutnik2022adaptation} (Fig.~\ref{fig:robot}), localization and mapping, motion planing, and image dehazing~\cite{drews2016underwater}. As such inferring depth from vision systems has been widely investigated in the last years. There is a range of sensors and imaging setups that can provide depth, such as stereo, multiple-view, and time-of-flight (ToF)~\cite{godard2017unsupervised,godard2019digging,ma2018sparse}. Monocular depth estimation is different from other vision systems in the sense that it uses a single RGB image with no special setup or hardware, and as such has many advantages. 
Because of mechanical design considerations, in many AUVs it is difficult to place a stereo setup with a baseline that is wide enough, so there monocular depth is particularly attractive and can be combined with other sensors (e.g., Sonars) to set the scale.

\begin{figure*}[t]
\centerline{\includegraphics[width=\textwidth,height=\textheight,keepaspectratio]{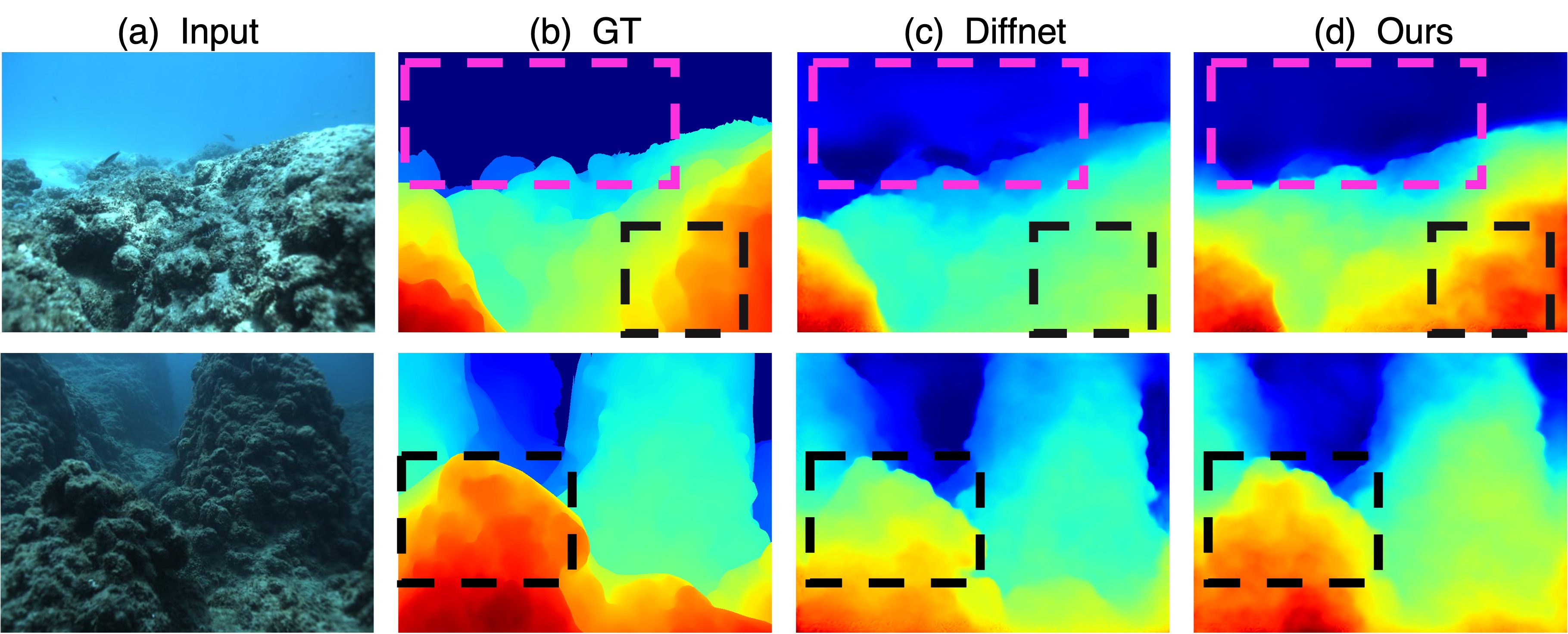}}
\caption{Example results on two underwater scenes from the \FLC dataset~\cite{FLSea}. a)~Input scene, b)~Ground truth, c)~Result of Diffnet~\cite{zhou2021self} and d)~our estimated depth map. Magenta rectangle marks background area where our method significantly improves the results, and black rectangles mark foreground objects with better estimation using our method.}
\label{fig:results_summary} 
\end{figure*}

Monocular depth methods can be trained either supervised or self-supervised. 
Naturally, supervised methods achieve higher accuracies, however, rely on having a substantial dataset with pairs of images and their ground-truth depth. This is very difficult to achieve underwater as traditional multiple-view methods struggle with appearance changes and are less stable. Additionally, optical properties of water~\cite{bekerman2020unveiling} change temporally and spatially, significantly changing scene appearance. Thus, for training supervised methods, a ground-truth dataset is needed for every environment, which is very laborious. Therefore, we chose to develop a self-supervised method, that requires only a set of consecutive frames for training. 

When testing state-of-the-art monocular depth estimation methods to underwater, new problems arise. Visual cues that one can benefit from above water might cause exactly the opposite and lead to estimation errors. 
Handling underwater scenes requires adding more constraints and using priors. Understanding the physical characteristics of underwater images can assist us in revealing new cues and using them for extracting depth cues from the images.

We improve self-supervised underwater depth estimation with the following contributions: \textbf{1)}~Examining how the reprojection loss changes underwater, \textbf{2)}~Handling background areas, \textbf{3)}~Adding a photometric prior, \textbf{4)}~Data augmentation specific for underwater.
To that end, we employ the \FLC dataset, published in~\cite{FLSea}. 

\section{Related Work}

\subsection{Supervised Monocular Depth Estimation}

In the supervised monocular depth task a deep network is trained to infer depth from an RGB image using a dataset of paired images with their ground-truth (GT) depth~\cite{eigen2014depth, liu2015deep}. 
Reference ground truth can be achieved from a depth sensor or can be generated by classic computer vision methods such as structure from motion (SFM) and from stereo. 
Li et. al~\cite{li2018megadepth} suggest to collect the training data by applying SFM on multi-view internet photo collections. Their network architecture is based on an hourglass network structure with suitable loss functions for fine details reconstruction in the depth map. a newer method~\cite{ranftl2021vision,bhat2021adabins} use transformers to improve performance.

\subsection{Self-Supervised Monocular Depth Estimation}

To overcome the hurdle of ground-truth data collection, it was suggested~\cite{godard2019digging, zhou2017unsupervised} to use sequential frames for self-supervised training leveraging the fact that they image the same scene from different poses. The network estimates both the  depth and the motion  between frames. The estimated camera motion between sequential frames constrains the depth network to predict  up to scale depth, and the estimated depth constrains the odometry network to predict  relative camera pose. The loss is the photometric reprojection error between two subsequent frames using the estimated depth and motion.

Monodepth2~\cite{godard2019digging} proposed to overcome occlusion artifacts  by taking the minimum error between preceding and following frames. 
DiffNet~\cite{zhou2021self} is based on  monodepth2~\cite{godard2019digging}  with two major differences. They replace the ResNet~\cite{he2016deep} encoder with  high-resolution representations using HRNet~\cite{wang2020deep} which was argued to perform better and added attention modules to the decoder. 
DiffNet~\cite{zhou2021self} is the current SOTA method on KITTI 2015 stereo dataset~\cite{geiger2012we}, the top benchmark for self-supervised monocular depth and also performed the best on our underwater images. Therefore, we base our work on it.

\subsection{Underwater Depth Estimation}\label{sec:uw_photometry}

Underwater, photometric cues have  been used for inferring depth from single images, as in scattering media the appearance of objects depends on their distance from the camera. Based on that several priors have been suggested for simultaneously estimating depth and restoring scene appearance.


One line of work is based on the dark channel prior (DCP)~\cite{he2010single} and several underwater variants UDCP~\cite{drews2013transmission, emberton2018underwater}, and the red channel prior~\cite{galdran2015automatic}. 
Some methods use the per-patch difference between the red channel and the maximum between the blue and the green as a proxy for distance, termed the maximum intensity prior (MIP) by Carlevaris-Bianco et al.~\cite{carlevaris2010initial}.
Song et al.~\cite{song2018rapid} suggested the underwater light attenuation prior (ULAP) that assumes the object distance is linearly related to the difference between the red channel and the maximum blue-green. The blurriness prior~\cite{peng2015single} leverages the fact that images become blurrier with distance. Peng and Cosman~\cite{peng2017underwater} combined this prior with MIP and suggested the image blurring and light absorption (IBLA) prior. Bekerman et al.~\cite{bekerman2020unveiling} showed that improving estimation of the scene's optical properties improves depth estimation. 

There have been also attempts of unsupervised learning-based underwater depth estimation.  UW-Net~\cite{gupta2019unsupervised} uses generative adversarial training  by learning the mapping functions between unpaired RGB-D terrestrial images and arbitrary underwater images. UW-GAN~\cite{hambarde2021uw} also used a GAN to generate depth, using supervision from a synthetic underwater dataset.  
These and others based the training on single images and none  uses geometric cues between subsequent frames for self-supervision as we do. As we show in the results, the self-supervision significantly improves the results.

\section{Scientific Background}

\subsection{Reprojection Loss}

The reprojection loss is the key self-supervision loss. 
It uses two sequential frames $[I_{t-1},I_{t}]$, where $t$ is the time index, together with the estimated extrinsic rotation, translation, and $\widehat{D}_{t}$, the estimated depth of frame $I_{t}$. These are used to compute the   coordinates  $\widehat{p}_{t-1}$ in $I_{t-1}$ that are the projection of the coordinates $p_{t}$ in $I_{t-1}$~\cite{zhou2017unsupervised}:
\begin{equation}
\widehat{p}_{t-1} \sim K\widehat{T}_{t\rightarrow t-1}\widehat{D}_{t}(p_{t})K^{-1}p_{t} \;\;.
\end{equation}
Here $\widehat{T}_{t\rightarrow t-1}$ is the inverse transform calculated from the extrinsic parameters and $K$ is  the camera intrinsic matrix, known from calibration. 
Then each pixel in the reprojected image $\widehat{I}_{t}(p_{t})$ is populated with values of ${I_{t-1}}(\widehat{p}_{t-1})$.


Based on color constancy the reprojection $\widehat{I}_{t}$ is supposed to be similar to the original frame ${I}_{t}$. 
Following~\cite{godard2019digging}, often the re-projection loss is a combination of two similarity measures, $L_1$ and
single scale structural similarity (SSIM)~\cite{wang2004image}:
\begin{equation}\label{eq:reprojloss}
L_{\rm reproj} = \alpha L_1(I_{t}, \widehat{I_{t}})+(1-\alpha)SSIM(I_{t}, \widehat{I_{t}})\;\;,
\end{equation}
weighted by $\alpha$,  commonly set to $=0.15$~\cite{godard2019digging}.

\subsection{Underwater Photometry}

As described in~\cite{bekerman2020unveiling}, the image formation model of a scene pixel $x$   in a participating medium such as underwater is composed of two additive components: 
\begin{equation}
I(x) = J(x)t(x) + A(1-t(x))~~,~~t=e^{-\chi d}\;\;.
\label{eq:I_uw}
\end{equation}
The scene radiance $J$ is attenuated by the medium. The medium transmission $t$ is exponential is the scene depth $d$ and and $\chi$, the medium's attenuation coefficient. 
Backscatter $A(1-t)$ is an additive component that stems from scattering along the line of sight, where $A$ is the global light in the scene. 

It is important to note that $\chi$ is wavelength dependant, i.e.,  each color channel attenuates differently with distance from the camera. In most water types the attenuation of red and near-infrared portions in water is much higher than the shorter visible wavelengths~\cite{pope1997absorption}. Hence, in underwater scenes, the red channel decreases faster with the distance. Based on this observation the ULAP prior was suggested~\cite{song2018rapid}.  It is calculated as the difference between the maximum value of $B$ and $G$, the blue and green color channels, and the value of $R$, the red color channel
\begin{equation}
u = \max(B,G) - R\;\;.
\label{eq:ULAP}
\end{equation}
According to~\cite{song2018rapid} the ULAP depth prior $u$ is supposed to be linearly related to the scene depth.

\section{Underwater self-supervised monocular depth estimation framework}

\subsection{Reprojection Loss Underwater}

\begin{figure}[t]
\centerline{\includegraphics[width=\linewidth]{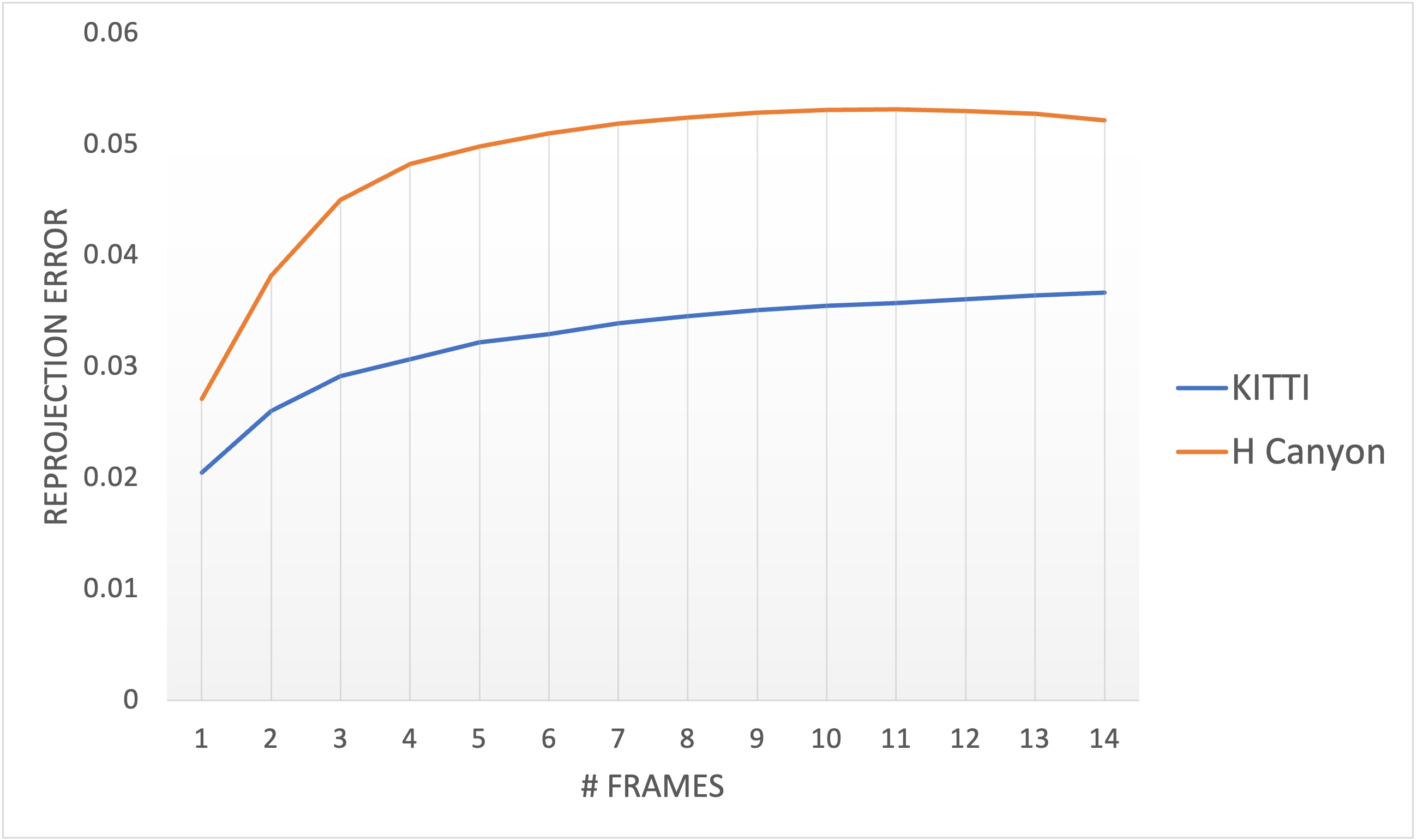}}
\caption{Reprojection loss error as a function of frame gap in KITTI and an \FLC subset (Horse Canyon). As expected, when the gap between frames increases, the error increases as well. This happens in both datasets but is more prominent underwater due to the effect of the medium.}\vspace{-0.3cm}
\label{fig:reprojLoss}
\end{figure}

\begin{figure}[t]\vspace{-0.1cm}
\centerline{\includegraphics[width=\linewidth]{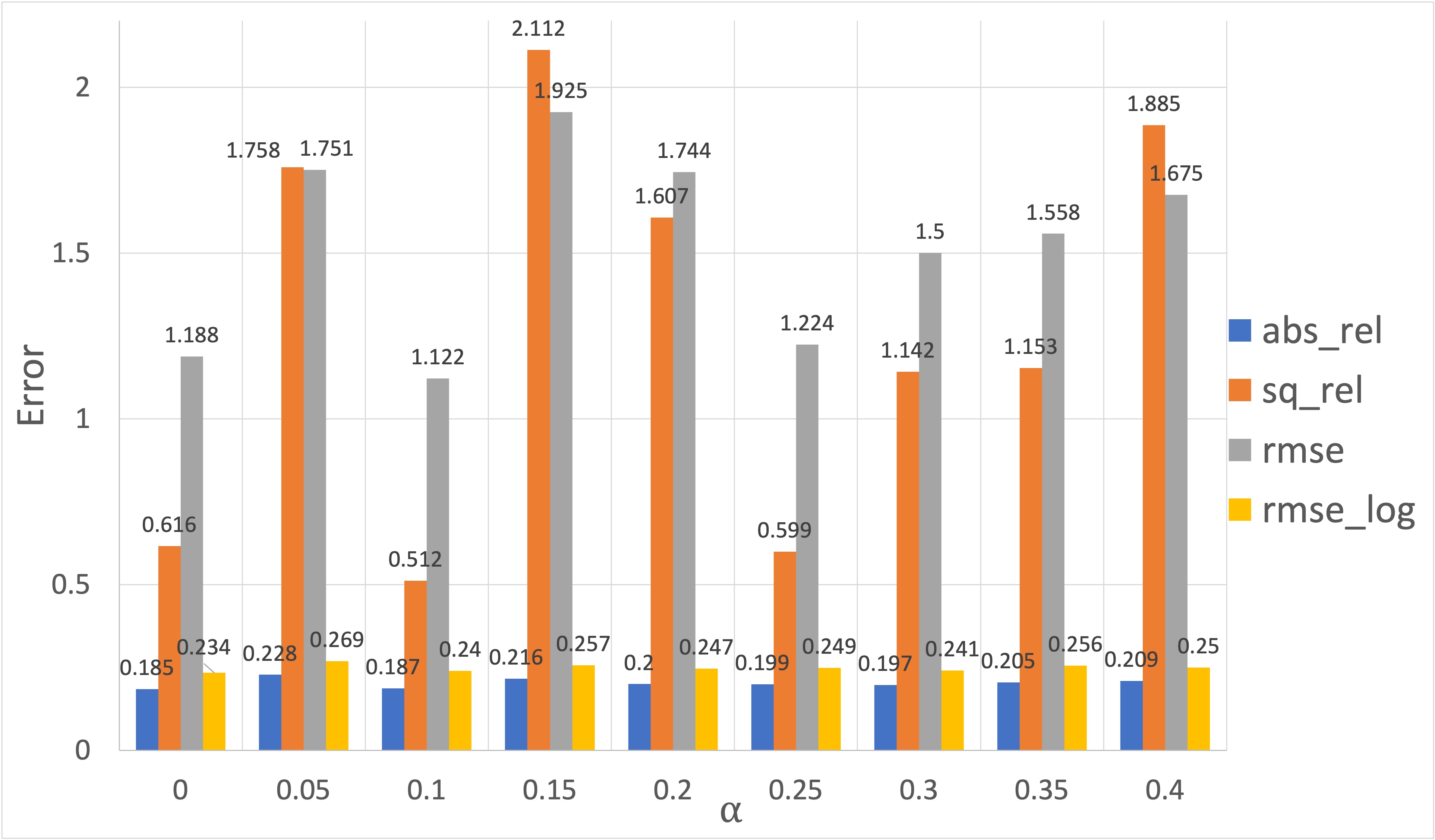}}
\caption{Reprojection loss error as a function of $\alpha$.} \vspace{-0.2cm}
\label{fig:alpha_exp}
\end{figure}

Following~\eqref{eq:I_uw} the participating medium greatly affects the acquired underwater images as a function of object depth. Hence,  camera movement underwater might lead to a significant difference in the captured subsequent images, questioning the validity of the reprojection loss~\eqref{eq:reprojloss} in this case. 
One solution to this is to insert the photometric model ~\eqref{eq:I_uw} into the loss function~\eqref{eq:reprojloss}. This would require estimation of additional parameters $\chi$ and $A$ and would add complexity. Before doing that, we conducted an experiment to examine the influence of the medium on the reprojection loss, as a function of inter-frame camera motion. Our assumption was that in nearby frames, the influence of the medium on the loss can be neglected.

Fig.~\ref{fig:reprojLoss} summarizes this analysis in comparison to a similar analysis on the KITTI dataset. 
The reprojection loss between subsequent frames in our test set is calculated using the predicted depth and camera poses. We repeat the same calculation for an increasing gap between the frames.  We see that in nearby subsequent frames the underwater loss is slightly larger than the outdoor error in KITTI, but is still very small. As expected, the error increases as the distance between subsequent frames grows. This points on the importance of high frame-rate imaging when acquiring training sets  underwater, and confirms our assumption that in our dataset the original loss can be used. 

The loss~\eqref{eq:reprojloss} that is commonly used combines $L_1$ which is a pixel-wise comparison, with SSIM that is a more general image quality measure with a weight of $\alpha=0.15$, i.e., SSIM receives a much larger weight. Since there are more illumination changes underwater we hypothesize that the ideal $\alpha$ value underwater should be lower. To test that, we conduct an experiment in which we run the baseline method with a range of $\alpha$ values. The results are summarized in Fig.~\ref{fig:alpha_exp}. We see that the both $\alpha=0$ and $\alpha=0.1$ result in lower errors, with a small preference for $\alpha=0.1$, which we choose to use in our experiments.


\subsection{Inferring Range in Areas Without Objects}\label{sec:LVW}

\begin{figure}[t]
 \centerline{\includegraphics[width=\linewidth]{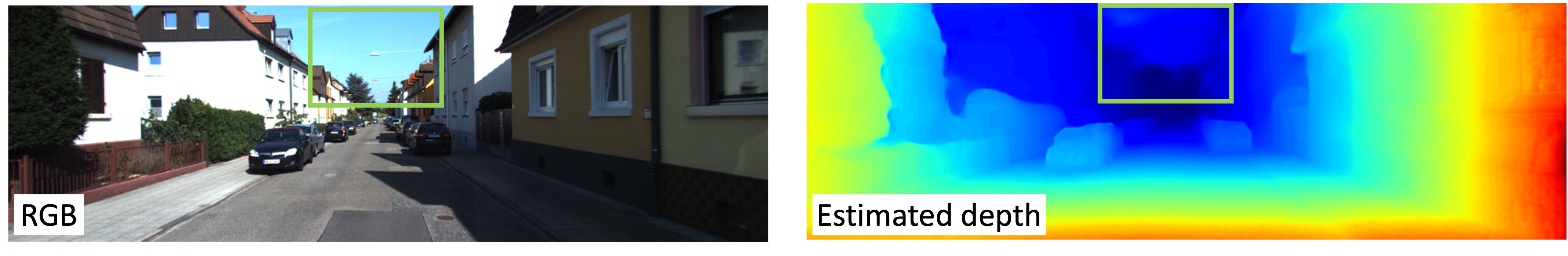}}
\caption{DiffNet depth prediction results on KITTI. There is no ground-truth for sky regions so the error there is not measured. We see that the depth of the sky in the image (green rectangle) is mistakenly predicted to be closer than the trees below it.}
\label{fig:kitti_sky}\vspace{-0.2cm}
\end{figure}
 
The re-projection loss \eqref{eq:reprojloss} minimizes the misalignment of details in the image. This creates an issue 
when estimating image areas that have no objects (e.g., sky, water background), since in textureless areas any depth results in a low reprojection loss. In KITTI, there is no ground-truth available for the sky as the measurements are LIDAR measurements that only reflect from nearby objects. However, when observing the results qualitatively, it is noticeable that some areas in the sky receive erroneous nearby ranges (see Fig.~\ref{fig:kitti_sky}). When using the depth inference to guide driving vehicles this is probably not an issue as the vehicles drive on the ground level and values in areas vertically above the car height are less relevant.

\begin{figure}[t]
 \centerline{\includegraphics[width=\linewidth]{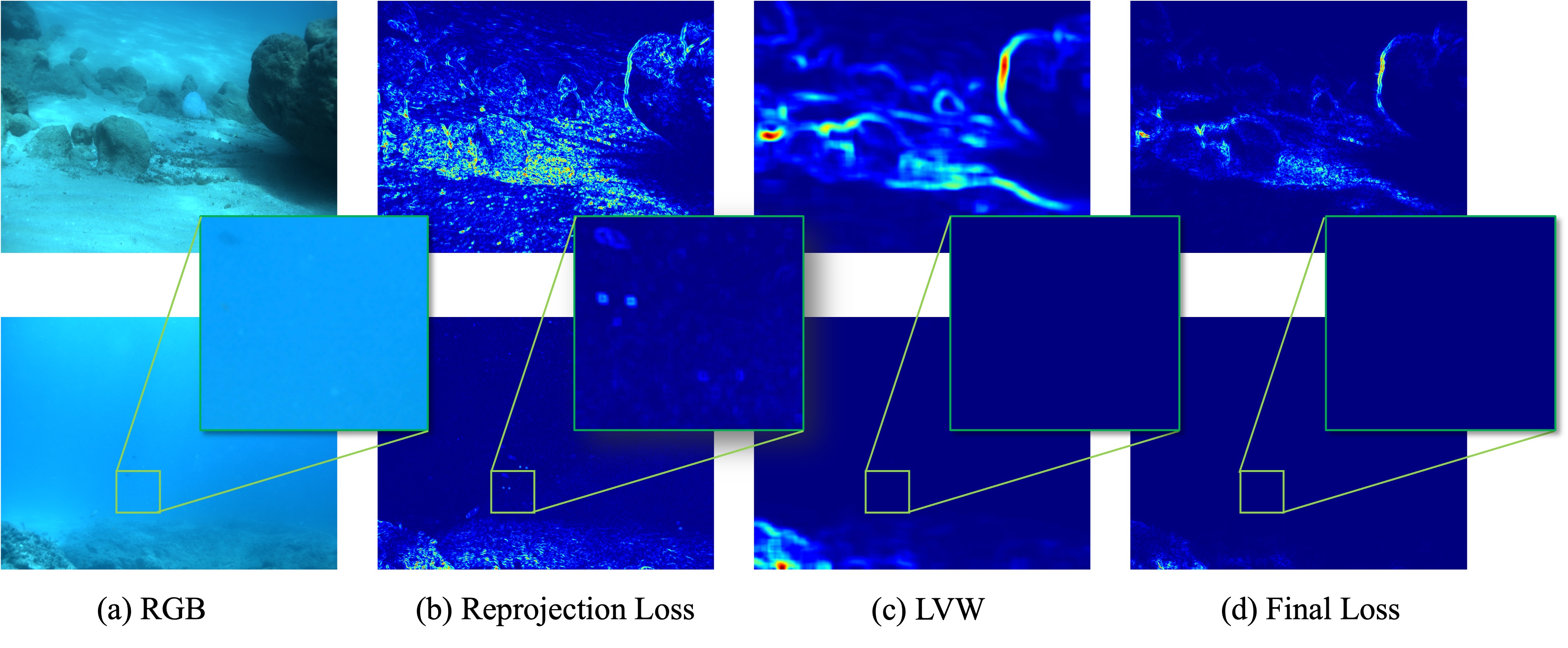}}
\caption{The effect of local variation map on the reprojection loss. a)~An underwater scene. b)~The reprojection loss~\eqref{eq:reprojloss} calculated between consecutive frames. (c)~The normalized LVW map ~\eqref{eq:lvnorm}. d)~The final loss~\eqref{eq:finallvw} after multiplication with the normalized LVW. In the original loss (b) reflections and non-uniform illumination introduce errors. The normalized LVW filters out this noise and leaves the real errors of the projection miss-alignment.}\vspace{-0.4cm}
\label{fig:lvw}
\end{figure}

However, underwater, vehicles regularly move vertically in a 3D space and require accurate range estimation also in areas that are vertically above them. If an object-less background area is mistakenly assigned a nearby value, it might affect the vehicle motion planning and the vehicle will attempt to bypass it without any reason. Moreover, this issue becomes more severe in underwater scenes, as ambient illumination is non-uniform and the background appearance can change between frames, increasing the reprojection error (e.g., the background noise in Fig.~\ref{fig:lvw}b). 
 Thus, this issue becomes critical underwater and we attempt to overcome it.



We want the loss to focus on the visible objects, such that it does not try to explain illumination changes in the object-less areas. For that, we propose the Local Variation Weight (LVW) mask $\sigma_k$. We calculate a local variation map over the image~\eqref{eq:lv}, which extracts interest areas in the image 
\begin{equation}\label{eq:lv}
\sigma_{k} = \mathbb{E}(x^2_{k}) - \mathbb{E}(x_{k}^2)\;\;,
\end{equation}
where $\mathbb{E}$ is the expectation operator and $x_{k}$ is an image region of size $k=25$.
This map is normalized between 0 and 1:
\begin{equation}\label{eq:lvnorm}
\hat{\sigma}_{k} = \frac{\sigma_{k} - \min(\sigma_{k})}{ \max(\sigma_{k}) - \min(\sigma_{k})}\;\;.
\end{equation}
and is used as weights on  the original re-projection loss~\eqref{eq:reprojloss}  to yield the final re-projection loss $\widehat{L}_{\rm reproj}$.
\begin{equation}\label{eq:finallvw}
\widehat{L}_{\rm reproj} = L_{\rm reproj} \cdot \hat{\sigma}_{k}\;\;.
\end{equation}
 A similar mask was used in~\cite{wang2011image} for image segmentation in noisy and textured environments. Fig.~\ref{fig:lvw} demonstrates the effect of LVW on two scenes. The LVW mask reduces some of the effects of flickering, backscatter and changing appearance of the rocks due to the combination of different camera orientation and nonuniform illumination.

\subsection{Underwater Light Attenuation Prior (ULAP)}

\begin{figure}[t]
\centerline{\includegraphics[width=0.8\linewidth]{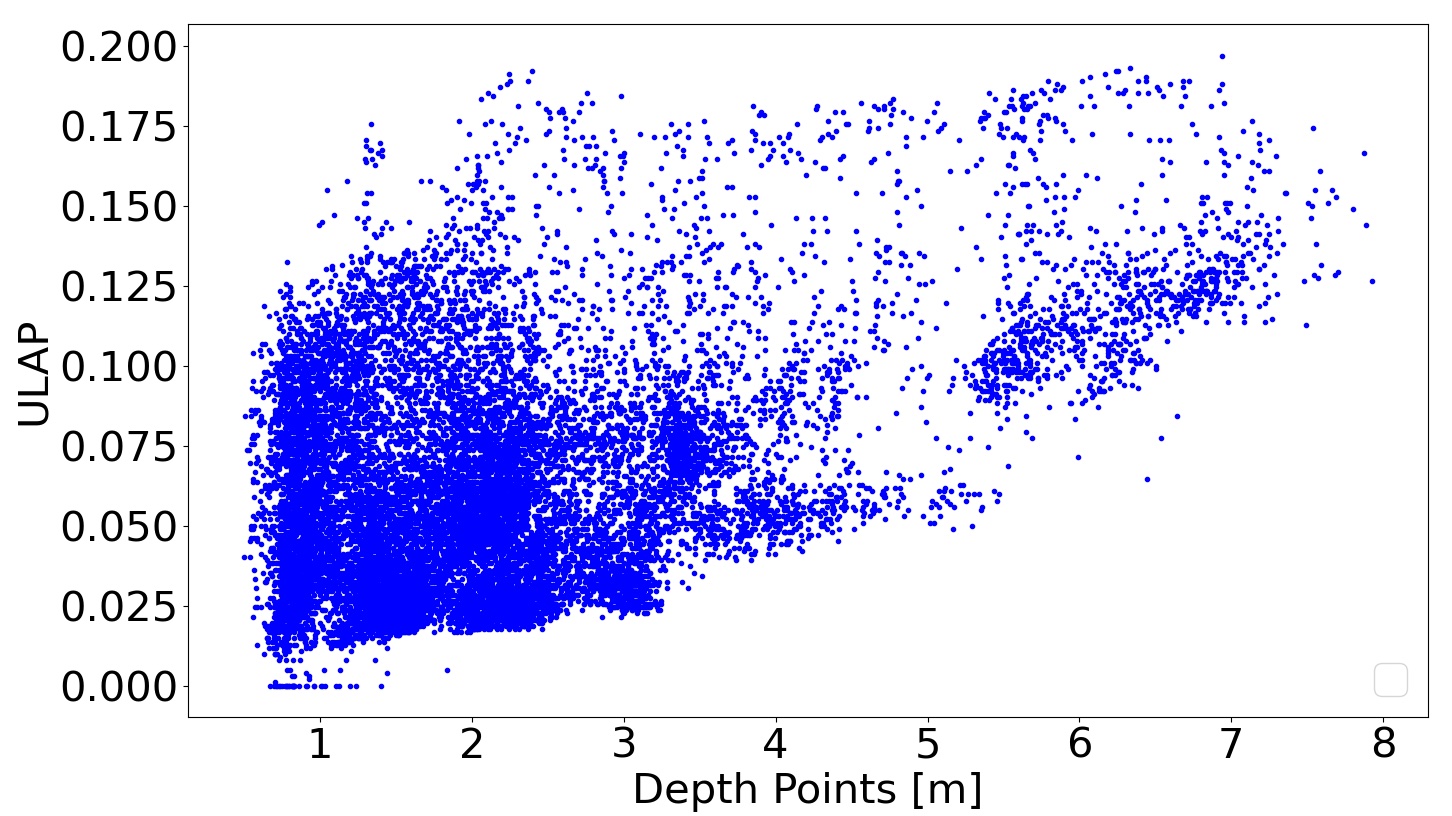}}
\caption{Correlation of the depth prior ULAP values versus ground truth depth. Pearson Coefficient between ground truth to ULAP equals 0.46.}
\label{fig:corrLoss}
\end{figure}

As discussed in Sec.~\ref{sec:uw_photometry}, underwater, photometric cues can aid depth estimation. Here we add the ULAP prior as guidance for the estimation. First, we examine the validity of the prior. 
In Fig.~\ref{fig:corrLoss} we show the correlation between both the ground truth depth with the ULAP~\eqref{eq:ULAP} calculated on our test set images. The correlation is $0.46$, which shows some relation but means ULAP by itself cannot be used for depth estimation.  
Using this insight, we encourage the correlation between the ULAP prior $u$ and our depth estimation $d$ by penalizing scores that are smaller than $1$:
\begin{equation}\label{eq:pearsonloss}
L_{\rm corr} = 1 - \frac{\sum(d - \bar{d}) \cdot (u - \bar{u})} {\sqrt{\sum(d - \bar{d})^2 \cdot \sum(u - \bar{u})^2}}\;\;,
\end{equation}
where  $\bar{d}$ is the mean depth over the image, and $\bar{u}$ is the mean of $u$.  The weight for this loss was empirically set to $1{\rm e}^{-5}$.

\subsection{Underwater Data Augmentation}

Compared to above-water haze-free images, in which the sky is usually uniformly illuminated, the underwater medium introduces light scattering which is changed by distance from the camera, camera orientation and the direction of sun. This could greatly affect unsupervised depth estimation since we do not expect to see projection errors in fully aligned regions. To generalize the network to perform well under different illuminations, we use dedicated data augmentation in training, using homomorphic filtering.

\begin{figure}[t]
\centerline{\includegraphics[width=0.8\linewidth]{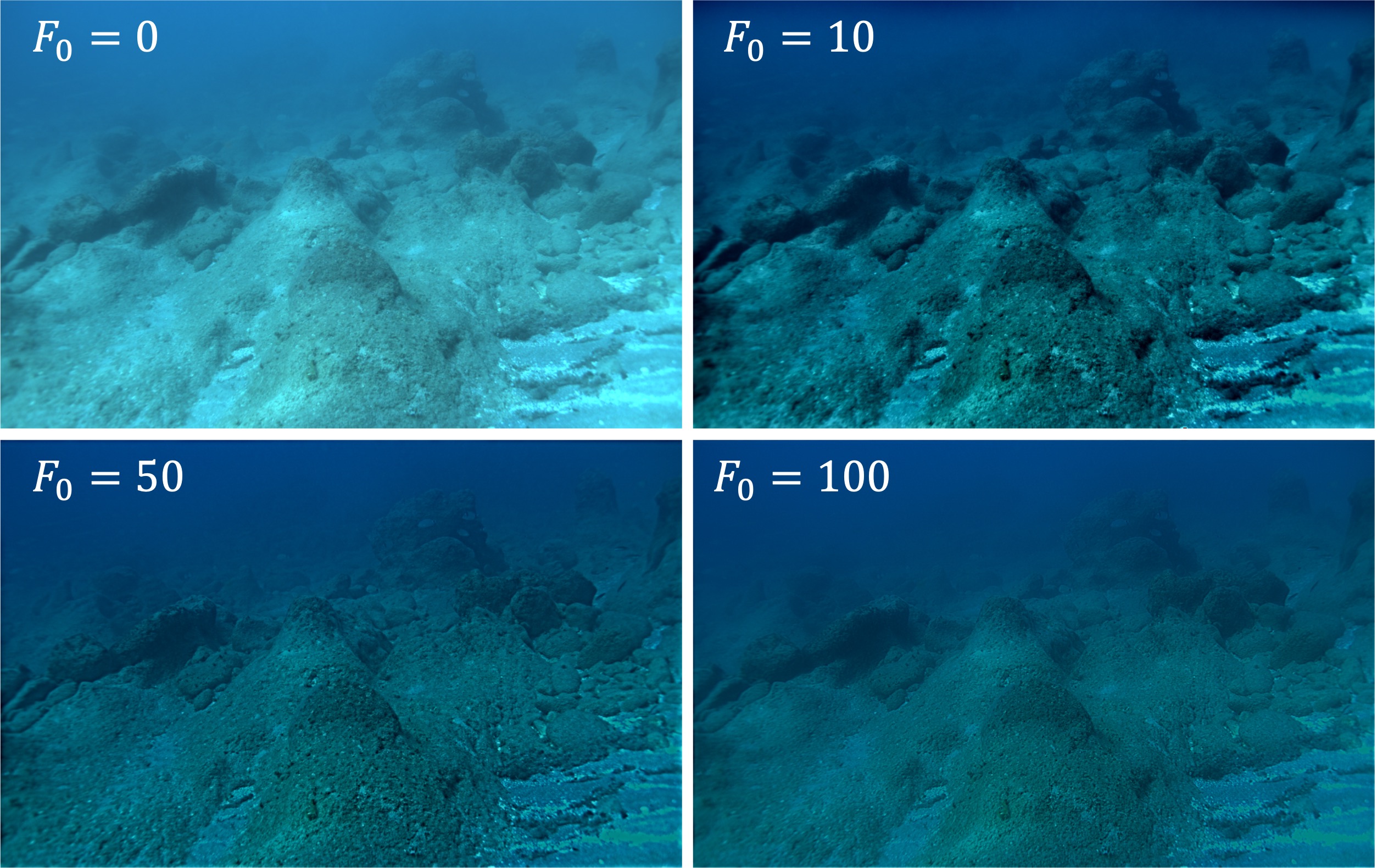}}
\caption{An example underwater image after homomorphic filtering with different cutoff frequency $D_0$ values~\eqref{eq:H}. $D_0$=0  is the original image, higher cuttoff frequencies eliminate low frequencies from the image and emphasize high frequencies.}
\label{fig:hf}
\end{figure}

Homomorphic filtering~\cite{adelmann1998butterworth} is an image processing filter that is used for image enhancement, denoising~\cite{gorgel2010wavelet} and non-uniform image illumination correction~\cite{kaeli2011improving}. The homomorphic filter serves as a high-pass filter, reducing low frequency variations that stem from  illumination changes, with a controllable cutoff frequency.  We use it to augment the input training images with a randomly parameterized homomorphic filter. Each input image goes through an homomorphic filter with a random uniformly distributed cuttoff frequency $F_0$ with values that range between $0$ to $250$. Setting $F_0=0$ yields the original image. This results in images with more homogeneous  illumination (see Fig.~\ref{fig:hf}) and aids training.

The homomorphic filter $H$ is a Butterworth high pass filter~\eqref{eq:H}, initialized with a cutoff frequency $F_0$  
\begin{equation}\label{eq:H}
H(z,w) = \left\{1 + \left[\frac{F_0}{F(z,w)}\right]^{2n}\right\}^{-1}\;\;,
\end{equation}
where $F(z,w)$ is 2D euclidean distance from point (z,w) to the center of the frequency space frame. We set $n$, the order of the filter, to be 2, which generates a moderate transition around the cutoff frequency. 
To apply the filter the RGB image is converted to YUV color space
$(y, u, v) = RGB2YUV(I)$ and the filter is applied in the Fourier space on the log of the $y$ color channel of the image:
\begin{equation}\label{eq:yhat}
\widehat{y} = \exp{\mathcal{F}^{-1}(HY)}\;\;, \;\; Y = \mathcal{F}(\log{y})\;\;.
\end{equation}
The filtered RGB image $\widehat{I}$  is reconstructed from $(\widehat{y}, u,v)$
\begin{equation}\label{eq:finalI}
\widehat{I} = YUV2RGB(\widehat{y}, u,v)\;\;.
\end{equation}

\section{Experiment Details}

\subsection{Training and Testing}

For evaluation we use the \FLC dataset~\cite{FLSea}. It contains 4 scenes: U Canyon, Horse Canyon, Tiny Canyon, Flatiron, consisting of 2901, 2444,1082, 2801 frames respectively. 
All scenes were acquired in the same region in the Mediterranean Sea. Ground truth depth and camera intrinsics were generated  using SFM (with the Agisoft software), and are known to contain some errors.

We split each one of the scenes into train (2751, 2651, 932, and 2444 respectively), evaluation (50) and test (last 150 frames of each of the scenes). Horse Canyon was used for training but excluded from the test set due to the apparent low quality of the ground truth. 
We trained the network using pretrained weights on KITTI as a start point, as this yielded better results than training from scratch.

\subsection{Background Error Estimation}

\begin{figure}[t]
\centerline{\includegraphics[width=\linewidth]{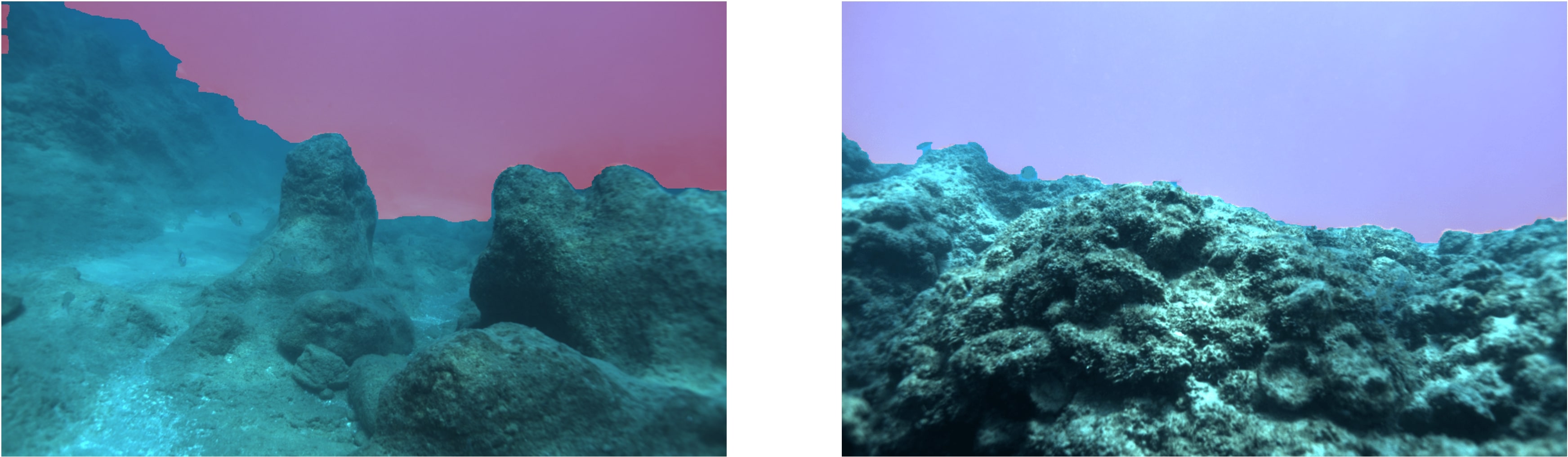}}
\caption{Examples of background masking of two underwater images taken from Tiny Canyon and flatiron.}
\label{fig:bgmasking}
\end{figure}

In most datasets, including ours, there is not ground-truth for background areas, as depth is measured only on objects. Thus,  performance is not  evaluated on background regions. Due to its importance on our case (Sec.~\ref{sec:LVW}), we specifically added a measure for the background error. 
The disparity in background areas is expected to be $0$, hence, we suggest an error measurement that penalizes pixels in the background that are greater than zero. Our motivation in this error calculation is to give lowest error to pixels with the lowest disparity $S$ estimation or alternatively farthest depth estimation
\begin{equation}\label{eq:bgerror}
BGerror = \frac{1}{m}\sum_{i=1}^{m}{S_{x\in{\rm bg}}}\;\;,
\end{equation}
where $m$ is the number of test images and $S$ is the predicted disparity map from the test set. 
For extracting the open water background ${\rm bg}$, we use the method described in~\cite{liba2020sky}, originally targeted for sky detection (see examples in Fig.~\ref{fig:bgmasking}).


\section{Results}


\begin{table*}[t]
\caption{An ablation analysis on the \FLC dataset. All methods perform better than the baseline. Augmentation contribute mostly to the background error reduction. Measurement reveals significant improvement in background depth estimation.}\vspace{-0.2cm}
\begin{center}
\begin{tabular}{|c|c|c|c|c|c|c|c|c|c|c|c|}
\hline
\textbf{$L_{\rm corr}$}& \textbf{\textit{LVW}}& \textbf{Augmentation}&
$\bf{\alpha}$&
\textbf{\textit{AbsRel}}& \textbf{\textit{SqRel}}& 
\textbf{\textit{RMSE}}& \textbf{\textit{RMSElog}}& \textbf{\textit{$\delta<1.25\uparrow$}}&
\textbf{\textit{$\delta<1.25^2\uparrow$}}& \textbf{\textit{$\delta<1.25^3\uparrow$}}& \textbf{\textit{BGerror}}\\
\hline

\multicolumn{3}{|c|}{\textbf{ U W - N E T }} & - &   0.527  &   1.765  &   1.725  &   1.961  &   0.337  &   0.565  &   0.699& 3.9247 \\
 \hline
\multicolumn{3}{|c|}{\textbf{ B A S E L I N E }} & 0.15 & 0.203&	1.955&	1.546&	0.245&	0.768&	0.923&	0.966&	1.381 \\
 \hline
 
 & & & 0.1 & 0.186&	1.828&	1.295&	0.222&	\colorbox{pink}{0.793}&	\colorbox{pink}{0.935}&	0.97&	1.396 \\
 
 \checkmark& & & 0.1 & 0.162&	0.245&	0.661&	0.209&	0.78&	0.934&	\colorbox{pink}{0.974}&	1.213 \\
 & \checkmark& & 0.1 & 0.158&	0.18&	0.644&	0.218&	0.768&	0.929&	0.963&	1.795 \\
 & & \checkmark & 0.1 & 0.18&	0.366&	0.775&	0.221&	0.751&	0.924&	0.97&	1.372 \\
 & \checkmark& \checkmark & 0.1 &0.165&	0.212&	0.7&	0.227&	0.774&	0.92&	0.958&	1.714\\
 \checkmark& & \checkmark& 0.1 &0.176&	0.399&	0.859&	0.213&	0.771&	0.928&	{0.974}&	1.098\\
 \checkmark& \checkmark& & 0.1 & \colorbox{pink}{0.156}&	\colorbox{pink}{0.146}&	0.589&	0.21&	0.775&	0.927&	0.972&	0.79 \\
 \checkmark& \checkmark& \checkmark & 0.1 &  0.158&	0.149&	\colorbox{pink}{0.581}&	\colorbox{pink}{0.208}&	0.778&	0.924&	0.969&	\colorbox{pink}{0.783}\\
\hline

\end{tabular}
\label{tableFLC}
\end{center}\vspace{-0.3cm}
\end{table*}

\begin{figure*}[t]
\centerline{\includegraphics[width=\textwidth,height=\textheight,keepaspectratio]{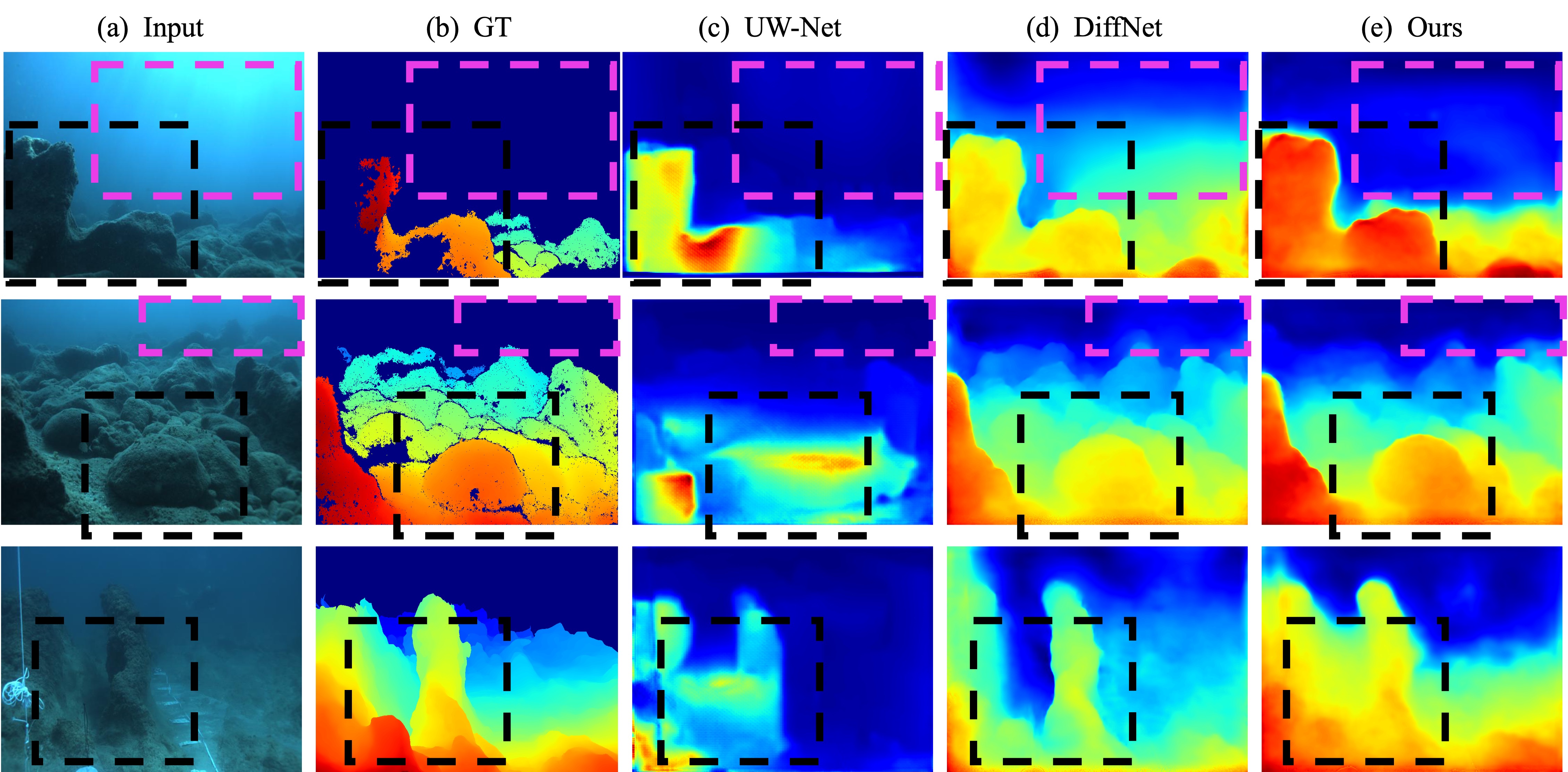}}
\caption{Example results on three underwater scenes from the \FLC dataset~\cite{FLSea}. a)~Input scene, b)~Ground truth, c)~UW-Net result, d)~Result of Diffnet and e)~our estimated depth map. Magenta rectangle marks background area where our method significantly improves the results, and Black rectangles mark foreground objects with better estimation using our method.}
\label{fig:comparisonSet} \vspace{-0.2cm}
\end{figure*}






Table~\ref{tableFLC} summarizes the results and the ablation study.
The results are reported using the evaluation metrics described in~\cite{eigen2014depth}.
Since we could achieve good background masking only on Tiny Canyon,  we calculate the background error only on this scene. 
Our method significantly improves the baseline DiffNet in all measures except of $\delta<1.25$ and $\delta<1.25^2$. These measures indicate the number of pixels with low errors. This means that our method is less accurate in fine depth estimation but more accurate in the global depth context, which is manifested in more accurate borders between objects and in correct depth decisions of objects with regard to other objects in the scene. Our method is also significantly better in the background error estimation in more than $30\%$. 
We see that even the baseline results of the above water method are  much better than the dedicated UWNET~\cite{gupta2019unsupervised}. The ablation study shows that using $\alpha=0.1$, $L_{\rm corr}$ and LVS always improve results. Augmentation with the homomorphic filter improves some of the measures and especially the background error.

\section{Discussion}

So far methods for monocular depth estimation underwater concentrated on leveraging photometric cues in single images, which is challenging to do in a self-supervised manner. We are the first to use self-supervision using subsequent frames, as successfully done above water. We show that using the standard above water SOTA methods underwater results in decent results but has room for improvement as it is not designed to cope specifically with appearance changes  caused by the medium. We analyze the performance of the standard reprojection loss and show that it can be used also underwater given the training set was acquired in high frame rate. We point to a problem that exists also above water in errors in estimating background areas that do not have ground truth. This was so far ignored above water, but in the three dimensional underwater realm it cannot be ignored and we suggest a weighed loss to mitigate this issue.

Since photometric priors on the single underwater images contain important information we combine one of them in the loss. In the future we plan to investigate how to further combine the single image information with the self-supervision obtained from subsequent frames. Lastly, we plan to incorporate this framework into a complete image restoration pipeline. Overall, our method  significantly improves the SOTA  in underwater monocular depth estimation and can substantially aid vision-based navigation and decision making in underwater autonomous vehicles.


{\small
\bibliographystyle{ieee_fullname}
\bibliography{refs}
}

\end{document}